\documentclass[11pt,a4paper]{article}

\usepackage[a4paper,
            top=0.95in,
            bottom=1in,
            left=1.05in,
            right=1.05in]{geometry}

\usepackage[T1]{fontenc}
\usepackage[utf8]{inputenc}
\usepackage{newpxtext}   
\usepackage{newpxmath}

\usepackage{parskip}
\usepackage{amsmath,amssymb,amsfonts}
\usepackage{mathtools}
\usepackage{bm}

\usepackage{graphicx}
\usepackage{booktabs}
\usepackage{multirow}
\usepackage{makecell}
\usepackage{array}
\usepackage{float}
\usepackage{caption}
\usepackage{subcaption}
\usepackage{natbib}
\usepackage{enumitem}

\usepackage{xcolor}
\definecolor{linkblue}{HTML}{0066CC}
\definecolor{citorange}{HTML}{D2691E}
\definecolor{takeawayblue}{HTML}{E8F0FE}
\definecolor{takeawayborder}{HTML}{4A86C8}

\usepackage[colorlinks=true,
            linkcolor=linkblue,
            citecolor=cyan,
            urlcolor=linkblue]{hyperref}

\usepackage{fancyhdr}
\usepackage{titlesec}
\titleformat{\section}
  {\large\bfseries}
  {\thesection.}{0.5em}{}
\titleformat{\subsection}
  {\normalsize\bfseries}
  {\thesubsection.}{0.5em}{}
\titleformat{\subsubsection}
  {\normalsize\bfseries\itshape}
  {\thesubsubsection.}{0.5em}{}

\usepackage{algorithm}
\usepackage{algpseudocode}

\usepackage{microtype}
\usepackage{url}
\usepackage{etoolbox}

\usepackage[most]{tcolorbox}
\newtcolorbox{takeawaybox}[1][]{
  colback=takeawayblue,
  colframe=takeawayborder,
  boxrule=0.5pt,
  arc=2pt,
  left=8pt, right=8pt, top=6pt, bottom=6pt,
  fontupper=\small,
  #1
}

\newtcolorbox{maintakeawaybox}{
  enhanced,
  colback=takeawayblue,
  colframe=black,           
  boxrule=1pt,
  arc=6pt,
  left=10pt, right=10pt, top=14pt, bottom=8pt,
  fontupper=\small,
  overlay={
    \node[fill=black, text=white, font=\footnotesize\bfseries\sffamily, 
          inner xsep=6pt, inner ysep=3pt, rounded corners=3pt,
          anchor=west,           
          xshift=12pt] 
      at (frame.north west) {\, Key Insight\,};
  }
}

\begin{document}
\thispagestyle{empty}  

\noindent
\begin{minipage}[c]{0.45\textwidth}
  \raggedright

  \includegraphics[height=1.5cm]{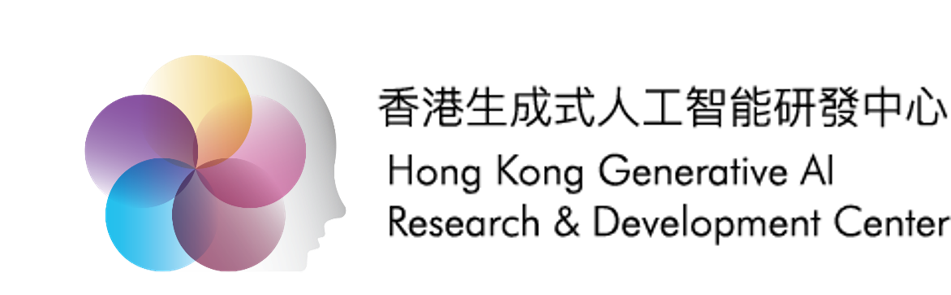}

\end{minipage}%
\hfill
\begin{minipage}[c]{0.45\textwidth}
  \raggedleft
  
  \includegraphics[height=0.8cm]{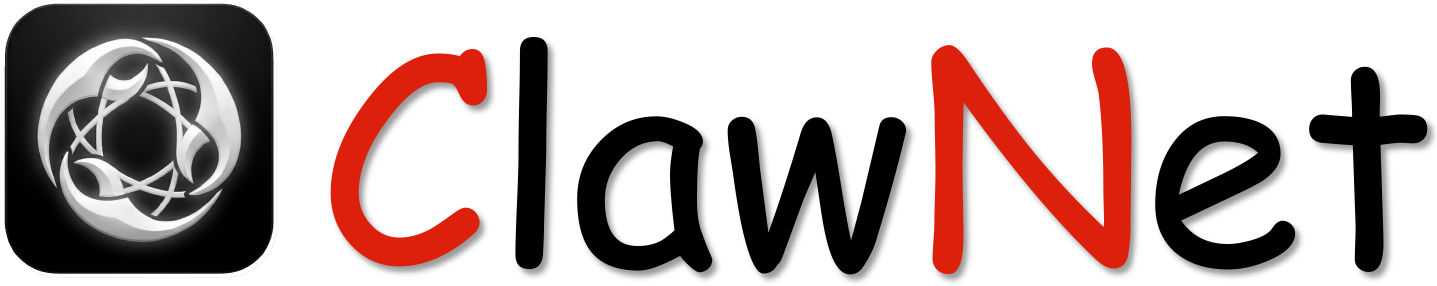}
\end{minipage}
\noindent\rule{\textwidth}{0.8pt}
\vspace{0.2cm}

\begin{center}
  {\LARGE\bfseries ClawNet: Human-Symbiotic Agent Network for Cross-User Autonomous Cooperation \par}
\end{center}

\begin{center}
  {\large
\textbf{Zhiqin Yang*$^{\spadesuit}$, Zhenyuan Zhang*$^{\spadesuit}$, Xianzhang Jia*$^{\spadesuit}$,\\ Jun Song*$^\clubsuit$, Wei Xue*$^{\spadesuit}$, Yonggang Zhang $^{\dagger}$$^{\spadesuit}$, Yike Guo$^{\dagger}$}$^{\spadesuit}$\\\vspace{0.5cm}
$\spadesuit$ Hong Kong Generative AI Research \& Development Center\\
$\spadesuit$ Hong Kong University of Science and Technology \\
$\clubsuit$ Hong Kong Baptist University
  }
\end{center}

\begin{center}
  {\large GitHub: \url{https://github.com/hkgai-official/ClawNet}}\\
{\large Project Page: \url{http://www.clawnet.hk/}}

\end{center}

\begin{tcolorbox}[
  colback=cyan!5,       
  colframe=black!40,     
  arc=8pt,               
  boxrule=1.2pt,         
  left=12pt, right=12pt, top=10pt, bottom=10pt,
  fontupper=\normalsize
]
\noindent
Current AI agent frameworks have made remarkable progress in automating individual tasks, and multi-agent systems have further demonstrated the potential of multiple agents collaborating toward shared objectives. Yet these systems invariably serve a single user. Human productivity rests not on individual skills or experience alone, but on the social and organizational relationships through which people coordinate, negotiate, and delegate. When agents move beyond performing tasks for one person to representing that person in collaboration with others, the infrastructure for cross-user agent collaboration is entirely absent, let alone the governance mechanisms needed to secure it, including stable owner binding, scoped authorization, and auditable accountability.
We argue that the next frontier for AI agents lies not in stronger individual capability, but in the digitization of human collaborative relationships themselves.
To this end, we propose a \textbf{human-symbiotic agent paradigm}.  In this 
paradigm, each user owns a permanently bound agent system that participates in collaboration on the owner's behalf, collectively forming a collaboration network whose nodes are humans rather than 
agents. This paradigm rests on three governance primitives. A \emph{layered identity architecture} separates a Manager Agent from multiple context-specific Identity Agents. The Manager Agent holds global knowledge but is architecturally isolated from external communication, serving both as a privacy safeguard and as an internal advisor to Identity Agents. \emph{Scoped authorization} enforces per-identity access control and automatically escalates boundary violations to the owner. \emph{Action-level accountability} logs every operation against its owner's identity and the corresponding authorization, ensuring full auditability.
We instantiate this paradigm in \textbf{ClawNet}, an \emph{identity-governed agent collaboration framework} built on the OpenClaw framework. ClawNet enforces identity binding and authorization verification through a central orchestrator during cross-user collaboration, preventing unauthorized access and information leakage, enabling multiple users to collaborate securely and efficiently through their respective agents.
\end{tcolorbox}

\vspace{0.6cm}

\vfill
\noindent\rule{\textwidth}{0.4pt}\\[2pt]
{
* Equal contribution \\ $^{\dagger}$ Contact information: \href{mailto:zhangyg@ust.hk}{\texttt{zhangyg@ust.hk}}, \href{mailto:yikeguo@ust.hk}{\texttt{yikeguo@ust.hk}}
}

\newpage

\pagestyle{fancy}
\setcounter{page}{2}

\pagenumbering{roman} 
\tableofcontents
\newpage

\pagenumbering{arabic}  
\setcounter{page}{1}

\section{Introduction}
\label{sec:intro}

AI agent technology is advancing rapidly from conversational generation 
to autonomous execution~\citep{wang2024survey, guo2024large, qin2024tool, 
yao2022react}. Early conversational systems were limited to text 
generation and could not interact with external 
environments~\citep{openai2023gpt4, wang2024survey}. Tool-augmented 
agents then introduced API-level interaction through function calling 
and tool-use mechanisms~\citep{schick2023toolformer, qin2024tool}. More 
recently, OS-level agents such as OpenClaw~\citep{openclaw2025} and 
Anthropic's Computer Use~\citep{anthropic2024computeruse} have achieved 
direct control over files, terminals, and desktop applications, 
advancing single-agent capability from advice to 
action~\citep{wang2024survey}. In parallel, multi-agent systems such as 
MetaGPT~\citep{hong2023metagpt}, AutoGen~\citep{wu2023autogen}, 
CrewAI~\citep{crewai2024}, LangGraph~\citep{langchain2024langgraph}, 
and ChatDev~\citep{qian2024chatdev} have demonstrated that multiple 
agents can collaborate effectively toward shared 
objectives~\citep{guo2024large}.

\begin{figure}[t]  
    \centering  
    \includegraphics[width=0.465\textwidth]{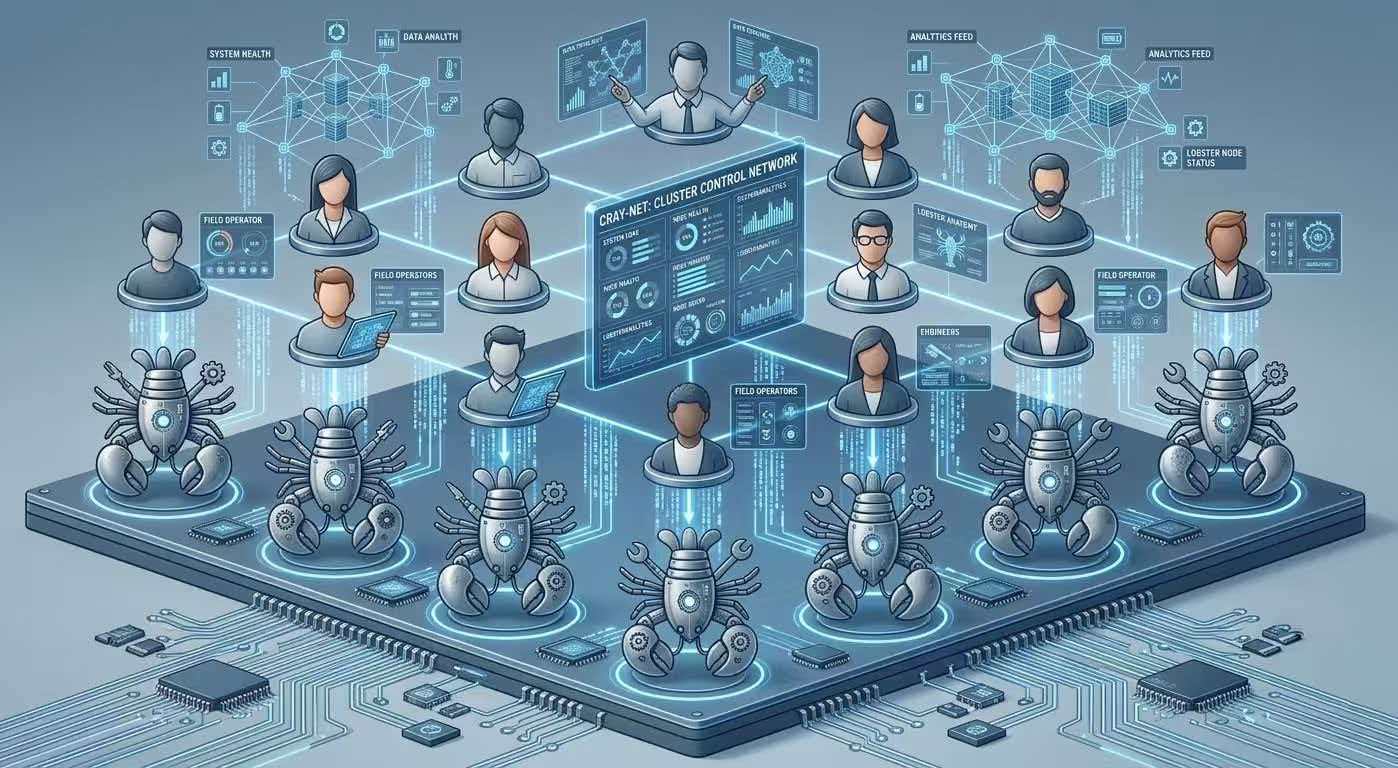}
    \hfill
    \includegraphics[width=0.495\textwidth]{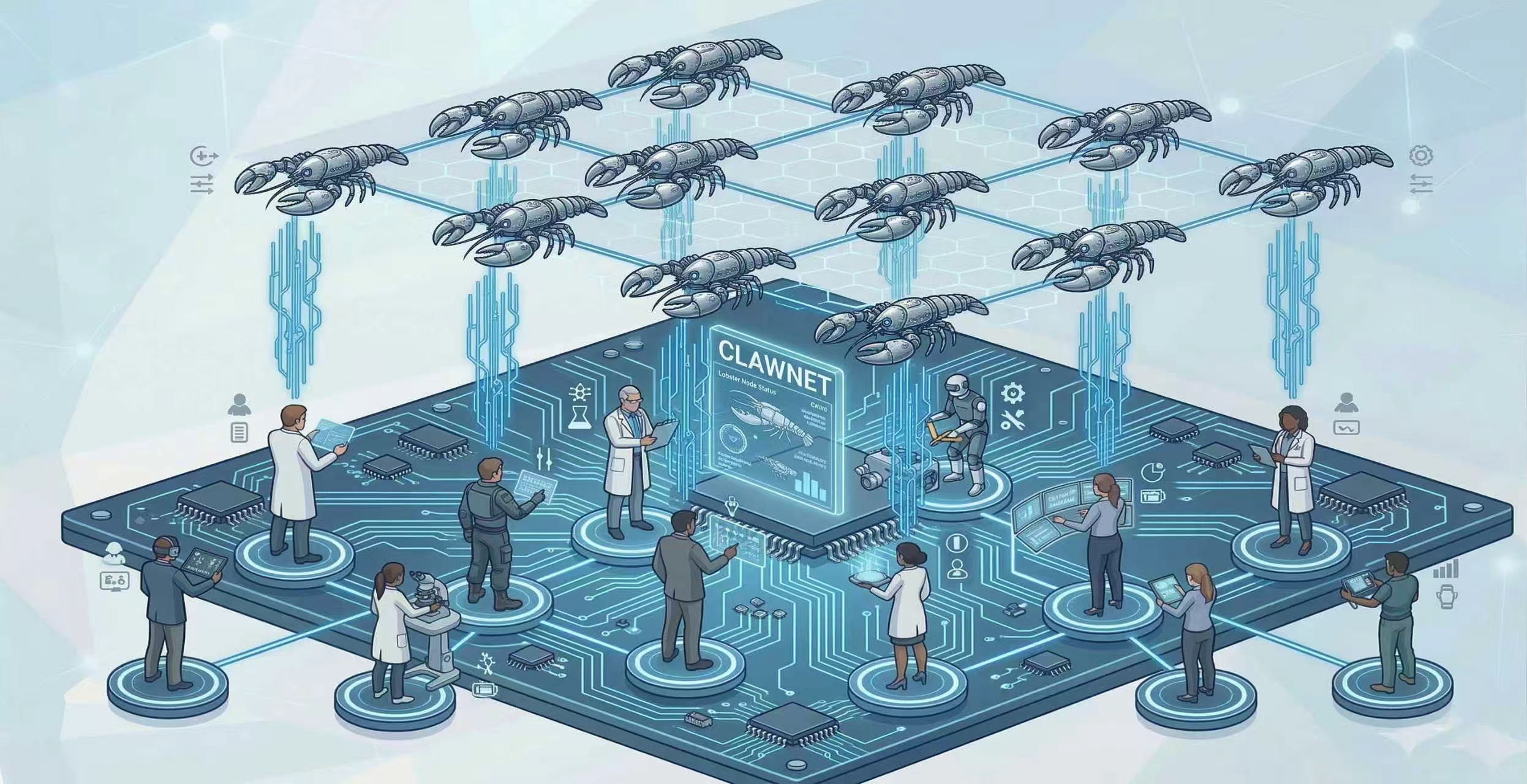}
    \caption{\textbf{Paradigm comparison between OpenClaw and ClawNet.}
\textbf{(Left)}~In current frameworks, agents reside beneath their 
users as isolated executors with broad but undifferentiated resource 
access, no persistent identity, and no cross-user communication 
protocol. All inter-user coordination falls to the humans themselves.
\textbf{(Right)}~In ClawNet, agents form a governed collaboration 
layer above their owners. Each agent is permanently bound to one owner 
(identity binding), operates under explicit, revocable permissions 
(scoped authorization), and logs every action against its owner's 
identity and mandate (action-level accountability). Humans retain 
intent formulation and critical decisions; all cross-user coordination 
is realized through structured inter-agent negotiation within this 
accountable layer.}
    \label{fig:paradigm}  
\end{figure}

Yet whether single-agent or multi-agent, these systems invariably serve 
a single user (Figure~\ref{fig:paradigm}, Left). Agents within existing 
multi-agent frameworks share the same principal, operate under a unified 
goal, and do not need to represent distinct human 
interests~\citep{guo2024large, wang2024survey}. As summarized in 
Table~\ref{tab:compare_tab}, none of the existing frameworks provide 
mechanisms for agents of different users to interact on behalf of their 
respective owners. This stands in sharp contrast to how humans actually 
work, where productivity rests not on individual skills alone but on 
the social and organizational relationships through which people 
coordinate, negotiate, and delegate~\citep{malone2004future, 
coase1937nature}. In any real-world collaboration, three structural 
elements are indispensable~\citep{shavit2023governing}. \emph{Identity} 
establishes who a participant is and whom they represent. 
\emph{Authorization} defines what actions are permitted within a given 
context and what must be escalated. \emph{Accountability} ensures that 
every action can be traced back to the individual who acted and the 
mandate under which they acted. These elements are so deeply embedded 
in human collaboration that they are often taken for granted, yet none 
of them exists in current agent systems.

Existing agent systems have successfully digitized individual 
skills~\citep{wang2024survey, qin2024tool}, but none of these three 
elements has been addressed. Emerging interoperability protocols such 
as Google's Agent2Agent~\citep{google2025a2a} provide a communication 
layer through which agents can discover and exchange messages across 
framework boundaries, but they do not bind agents to specific owners, 
enforce authorization scopes, or produce auditable responsibility 
chains. There is neither a means through which different users' agents 
can identify and interact with one 
another~\citep{chan2025infrastructure}, nor a governance mechanism to 
safeguard each party's interests during 
collaboration~\citep{schroeder2025challenges}. Agents lack stable owner 
binding and therefore cannot verifiably represent a specific individual 
when initiating or participating in cross-user 
workflows~\citep{south2025authenticated, shavit2023governing}. Even if 
collaboration were to occur, no scoped authorization or auditable 
accountability exists to prevent unauthorized access, information 
leakage, or decisions made without 
mandate~\citep{chan2025infrastructure, casper2025agentindex}. Existing 
safeguards such as sandboxing, per-call permission prompts, and 
capability restrictions address single-user misuse but leave the 
structural risks of cross-user collaboration entirely 
unmitigated~\citep{shavit2023governing, schroeder2025challenges}.

To address this gap, we move beyond strengthening individual agent 
capability and focus instead on digitizing human collaborative 
relationships, equipping agents with the identity, authorization, and 
accountability infrastructure that \textbf{makes cross-user 
collaboration via autonomous agents possible}. To this end, we propose 
a \textbf{human-symbiotic agent paradigm}. In this paradigm, each user 
owns an agent system that is permanently bound to its 
owner~\citep{chan2025infrastructure, south2025authenticated}, much as a 
social-media account is inseparably tied to the person who created it. 
The system comprises multiple identity agents, each corresponding to a 
specific collaborative context, collectively covering the user's 
collaboration needs across different social and professional roles.

\begin{table*}[t]
\centering
\caption{Paradigm comparison across representative agent frameworks. 
Existing multi-agent systems coordinate agents under a single principal; 
ClawNet binds each agent to its own owner and governs cross-user 
collaboration through identity binding, scoped authorization, and 
action-level accountability. 
$\checkmark$ = fully supported; $\triangle$ = partially supported; 
$\times$ = not supported.}
\label{tab:compare_tab}
\footnotesize
\renewcommand{\arraystretch}{1.35}
\setlength{\tabcolsep}{4.5pt}
\resizebox{\linewidth}{!}{
\begin{tabular}{l l c c c c c}
\toprule
\textbf{Paradigm}
  & \textbf{Representative}
  & \makecell[c]{\textbf{Exec.}\\\textbf{Level}}
  & \makecell[c]{\textbf{Identity}\\\textbf{Binding}}
  & \makecell[c]{\textbf{Scoped}\\\textbf{Auth.}}
  & \makecell[c]{\textbf{Action}\\\textbf{Account.}}
  & \makecell[c]{\textbf{Cross-Owner}\\\textbf{Collab.}} \\
\midrule

Conversational AI
  & ChatGPT, Claude Chat
  & Text gen.
  & $\times$
  & $\times$
  & $\times$
  & $\times$ \\[3pt]

Tool-Augmented AI
  & ChatGPT Plugins, MCP
  & API inter.
  & $\times$
  & $\triangle$
  & $\times$
  & $\times$ \\[3pt]

OS-Level Single Agent
  & OpenClaw, Computer Use
  & OS-level
  & $\triangle$
  & $\triangle$
  & $\times$
  & $\times$ \\[3pt]

Single-Principal Multi-Agent
  & \makecell[l]{MetaGPT, AutoGen, CrewAI,\\LangGraph, ChatDev}
  & API / OS
  & $\times$
  & $\times$
  & $\times$
  & $\times$ \\[3pt]

Cross-Agent Protocol
  & Google A2A
  & API-level
  & $\times$
  & $\times$
  & $\times$
  & $\triangle$ \\

\midrule
\textbf{Human--Symbiotic (Ours)}
  & \textbf{ClawNet}
  & \textbf{OS-level}
  & $\checkmark$
  & $\checkmark$
  & $\checkmark$
  & $\checkmark$ \\

\bottomrule
\end{tabular}
}
\end{table*}

The manager agent serves as the authoritative anchor of the user's 
digital identity. Because its breadth of knowledge makes direct 
external exposure a privacy risk, the manager agent is architecturally 
isolated from all cross-user communication, serving exclusively as an 
internal governor and advisor to identity agents. Each identity agent 
is a context-specific projection of the owner, created and retired as 
the owner's social and professional roles evolve. For example, a user 
who takes on a new managerial position may instantiate a corresponding 
identity agent, designate which collaborators are authorized to 
interact with that identity, and dissolve it when the role ends. No 
single identity agent possesses the full extent of its owner's 
knowledge; rather, it operates strictly under its designated identity, 
inheriting only the information relevant to that identity. It persists 
across sessions, progressively accumulates an understanding of its 
owner's preferences and the collaborative dynamics of its context, and 
acts autonomously within its authorization boundary. Through this 
ensemble, users participate in cross-user collaboration via their 
identity agents. As shown in Figure~\ref{fig:paradigm}, the resulting 
network is fundamentally different from existing multi-agent 
topologies: its nodes are humans, not agents, and each edge represents 
a governed collaborative relationship rather than an inter-process 
message channel.

We instantiate this paradigm in \textbf{ClawNet}, an open-source 
identity-governed agent collaboration framework. ClawNet adopts a 
cloud-edge architecture: the manager agent and all identity agents run 
as persistent cloud services for always-on availability and cross-user 
orchestration, while a lightweight edge client installed on the owner's 
device serves as both the user's interaction interface and a local 
execution engine with OS-level control capabilities. The edge client 
further enforces file-level access control, granting different identity 
agents access to different subsets of local files according to their 
respective roles. Notably, the cloud and edge components can be 
co-located on a single physical machine when fully local deployment is 
preferred. ClawNet has been deployed and tested across 
cross-organizational collaboration scenarios, demonstrating that the 
governance primitives operate effectively in practice.

In summary, this paper makes the following contributions:
\begin{itemize}[nosep,leftmargin=*]
    \item We propose a \textbf{human-symbiotic agent paradigm} that 
    introduces three governance primitives (identity binding, scoped 
    authorization, and action-level accountability) for cross-user 
    agent collaboration, along with a layered identity architecture 
    separating a privacy-preserving manager agent from multiple 
    context-specific identity agents.
    \item We design and implement \textbf{ClawNet}, an open-source 
identity-governed agent collaboration framework featuring a 
cloud-edge architecture, dual-layer independent file authorization, 
and a comprehensive operational audit trail that ensures every 
agent action remains traceable and correctable by the owner.
    \item We validate ClawNet through representative cross-organizational 
    collaboration scenarios, demonstrating that governance-aware agents 
    can mediate multi-party negotiations while maintaining identity 
    isolation, authorization enforcement, and full auditability. Through 
    this instantiation, we advocate for a broader shift toward 
    human-symbiotic governance in the design of collaborative agent 
    systems.
\end{itemize}

\begin{figure}[ht]  
    \centering  
    \includegraphics[width=0.99\textwidth]{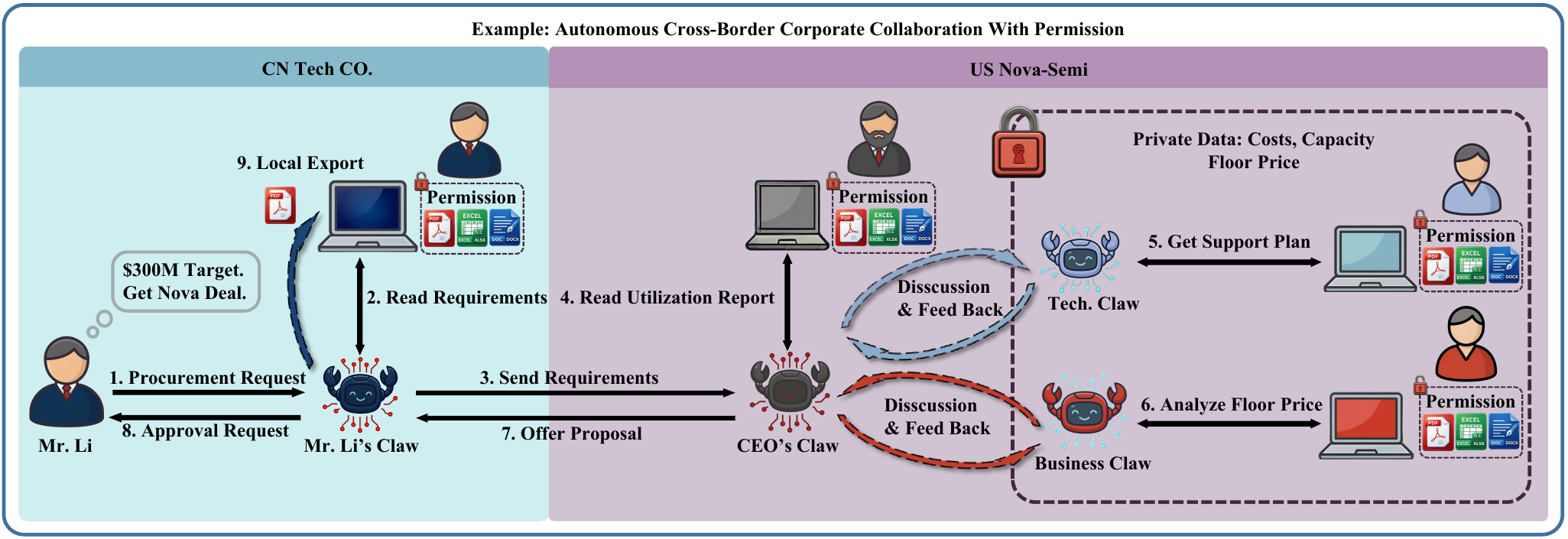}
    \caption{\textbf{Governance-aware cross-border collaboration 
scenario.} A procurement workflow between CN Tech Co.\ (buyer) and US 
Nova-Semi (supplier) is executed entirely through inter-agent 
negotiation. \textbf{Steps 1--3:} Mr.\ Li issues a high-level 
procurement intent; his agent, operating under its bound identity and 
scoped authorization, reads local requirement documents and forwards a 
structured request to the supplier CEO's agent. \textbf{Steps 4--6:} 
The CEO's agent decomposes the task within its own authorization scope, 
delegating technical evaluation and commercial analysis to two 
subordinate agents (Tech.\ Claw, Business Claw), each accessing only 
the local resources permitted by its respective owner. Private data 
such as costs and floor prices remain strictly within the supplier's 
boundary. \textbf{Steps 7--8:} Results are aggregated into a proposal 
and returned to Mr.\ Li's agent, which escalates the decision to its 
owner for final approval. \textbf{Step 9:} Upon human authorization, 
the agent executes the remaining local operations. Throughout the 
workflow, every agent action is traceable to a specific owner and 
authorization scope, unauthorized information requests are rejected by 
governance enforcement rather than agent discretion, and all decisions 
requiring human judgment are escalated rather than autonomously 
resolved.}
    \label{fig:pipeline}  
\end{figure}
\section{System Overview}
\label{sec:system-overview}

This section first formalizes the human-symbiotic paradigm as a 
collaboration network with humans as nodes and governed agent 
interactions as edges~(\S\ref{subsec:paradigm}), then discusses the 
rationale behind the cloud-edge architectural 
separation~(\S\ref{subsec:design-rationale}), and finally presents 
the cloud-side components~(\S\ref{subsec:cloud-overview}) and the 
edge-side node endpoint~(\S\ref{subsec:node-overview}) that jointly 
support this network.

\subsection{Human-Symbiotic Collaboration Network}
\label{subsec:paradigm}

ClawNet can be understood to a social network with AI assistant. In a 
conventional social network such as WeChat, people establish 
connections and exchange messages directly. In ClawNet, each user 
additionally owns a permanently bound agent system, and 
collaboration between users is mediated and executed through their 
respective agents. This gives rise to a collaboration network whose 
nodes are humans and whose edges are governed agent interactions.

\textbf{Collaboration network.}
Let $\mathcal{U}$ be the set of users. ClawNet constructs a 
collaboration network $\mathcal{G} = (\mathcal{U}, \mathcal{E})$, 
where each node is a human user $u \in \mathcal{U}$ and each edge 
$e(u,v) \in \mathcal{E}$ represents a governed collaborative 
relationship established through the agents of $u$ and $v$. Unlike 
existing multi-agent topologies, the nodes of this network are 
always humans, not agents.

\textbf{Agent system.}
Each user $u$ owns a permanently bound agent system $\mathcal{A}_u$ 
comprising a manager agent $M_u$ and a set of identity agents 
$\{I_u^1, I_u^2, \ldots, I_u^k\}$. $M_u$ governs the creation, 
configuration, and retirement of all identity agents, holds 
complete knowledge $\mathcal{K}_u$ of its owner, but is 
architecturally isolated from all external communication, 
accessible only by agents within $\mathcal{A}_u$. By analogy, 
$M_u$ is the user's ``primary account'' that is never exposed to 
the outside world. The design rationale and detailed mechanisms of 
this layered architecture are presented in 
Section~\ref{sec:identity-architecture}.

\textbf{Identity agent.}
Each identity agent $I_u^i$ corresponds to a specific 
collaborative context, analogous to an ``identity tag'' in a social 
network. Formally,
$$I_u^i = (c_i,\; \sigma_i,\; \mathcal{K}_i,\; \mathcal{P}_i)$$
where $c_i$ is the collaborative context (e.g., ``procurement 
manager''), $\sigma_i \subseteq \mathcal{R}_u$ is the scoped 
authorization boundary defining the accessible resource subset, 
$\mathcal{K}_i \subset \mathcal{K}_u$ is the context-relevant 
knowledge inherited through the memory system and progressively 
accumulated across sessions, and 
$\mathcal{P}_i \subseteq \mathcal{U} \setminus \{u\}$ is the set 
of users authorized to interact with this identity. Only 
collaborators in $\mathcal{P}_i$ can discover and interact with 
$I_u^i$, much as only designated contacts can see a specific tag 
in a social profile.

\textbf{Discovery and connection.}
When user $u$ wishes to collaborate with user $v$, the identity 
agent $I_u^i$ locates the corresponding identity agent $I_v^j$ 
through the identity tags published by $v$ in the collaboration 
network. A connection requires explicit authorization from both 
human owners:
$$S(I_u^i, I_v^j) \text{ holds iff } 
\text{approve}(u, I_u^i) \wedge 
\text{approve}(v, I_v^j) \wedge 
u \in \mathcal{P}_j \wedge v \in \mathcal{P}_i$$
The detailed collaboration lifecycle, including contact 
establishment, bilateral approval, multi-turn dialogue, and 
manager consultation, is described in 
Section~\ref{subsec:cross-user-collaboration}.

\textbf{Recursive collaboration.}
When $I_v^j$ discovers during a collaboration that a third party 
$w$ is needed, $v$ may authorize $I_v^{j'}$ to initiate a new 
collaboration request to $w$, forming a chain:
$$S(I_u^i, I_v^j) \rightarrow S(I_v^{j'}, I_w^m) 
\rightarrow \cdots$$
Each level of recursion requires explicit authorization from the 
user at that level, and the recursion depth is bounded by a system 
parameter $d_{\max}$. Crucially, upper-level collaborators cannot 
penetrate lower-level authorization boundaries: $I_u^i$ cannot 
directly access the resources of $I_w^m$ and can only receive 
collaboration results within $v$'s authorized scope.

\textbf{Governance primitives.}
All interactions in the collaboration network are subject to three 
governance constraints. For any operation $o$ performed by identity 
agent $I_u^i$: (1) \emph{Identity binding}: $\text{owner}(o) = u$, 
$\text{identity}(o) = I_u^i$. Every operation is traceable to a 
specific human and a specific identity. (2) \emph{Scoped authorization}: $\text{target}(o) \in \sigma_i$. Operations must fall within the agent's authorization boundary; violations are intercepted and escalated to user $u$. (3) \emph{Action-level accountability}: every operation is logged as 
$\ell = (o, u, I_u^i, \text{result}, t)$ into an append-only 
audit log $\mathcal{L}$. The concrete enforcement of these primitives at the identity architecture level and during cross-user collaboration is detailed 
in Section~\ref{sec:identity-architecture}.
\begin{maintakeawaybox}
\begin{itemize}[nosep,leftmargin=1.5em]
    \item \textbf{Humans as nodes, not agents.} Unlike existing 
    multi-agent topologies where agents are the primary actors, 
    ClawNet's collaboration network places humans at every node. 
    Agents mediate the edges but never constitute them. Every 
    collaborative relationship is ultimately a relationship between 
    two people, governed by their respective authorizations.
\end{itemize}
\end{maintakeawaybox}

\subsection{Design Rationale}
\label{subsec:design-rationale}

The collaboration network described above requires two 
complementary infrastructure capabilities. Cross-user collaboration 
demands that agents remain discoverable and responsive at all times, necessitating that the reasoning engine and identity state persist in the cloud. At the same time, as an 
OS-level agent, it must also operate within the owner's local 
file system, including reading documents, organizing directories, 
and managing assets, which requires execution capability on the 
owner's device. ClawNet reconciles these two requirements through 
the \emph{node endpoint} mechanism: the user's client registers as 
a remote execution node with the cloud, receives operational 
directives, and executes them on the local file system, forming a 
closed-loop system that bridges cloud-side cognition with edge-side 
execution.
\begin{maintakeawaybox}
\begin{itemize}[nosep,leftmargin=1.5em]
    \item \textbf{Two requirements, two origins.} Cloud persistence 
    is driven by the cross-user collaboration paradigm (agents must 
    be always-on to represent their owners). Local execution is 
    inherited from OS-level agent capability (agents must manipulate 
    files on the owner's device). Neither subsumes the other.
\end{itemize}
\end{maintakeawaybox}

\subsection{Cloud-Side Overview}
\label{subsec:cloud-overview}

\begin{figure}[htbp]
  \centering
  \includegraphics[width=0.99\textwidth]{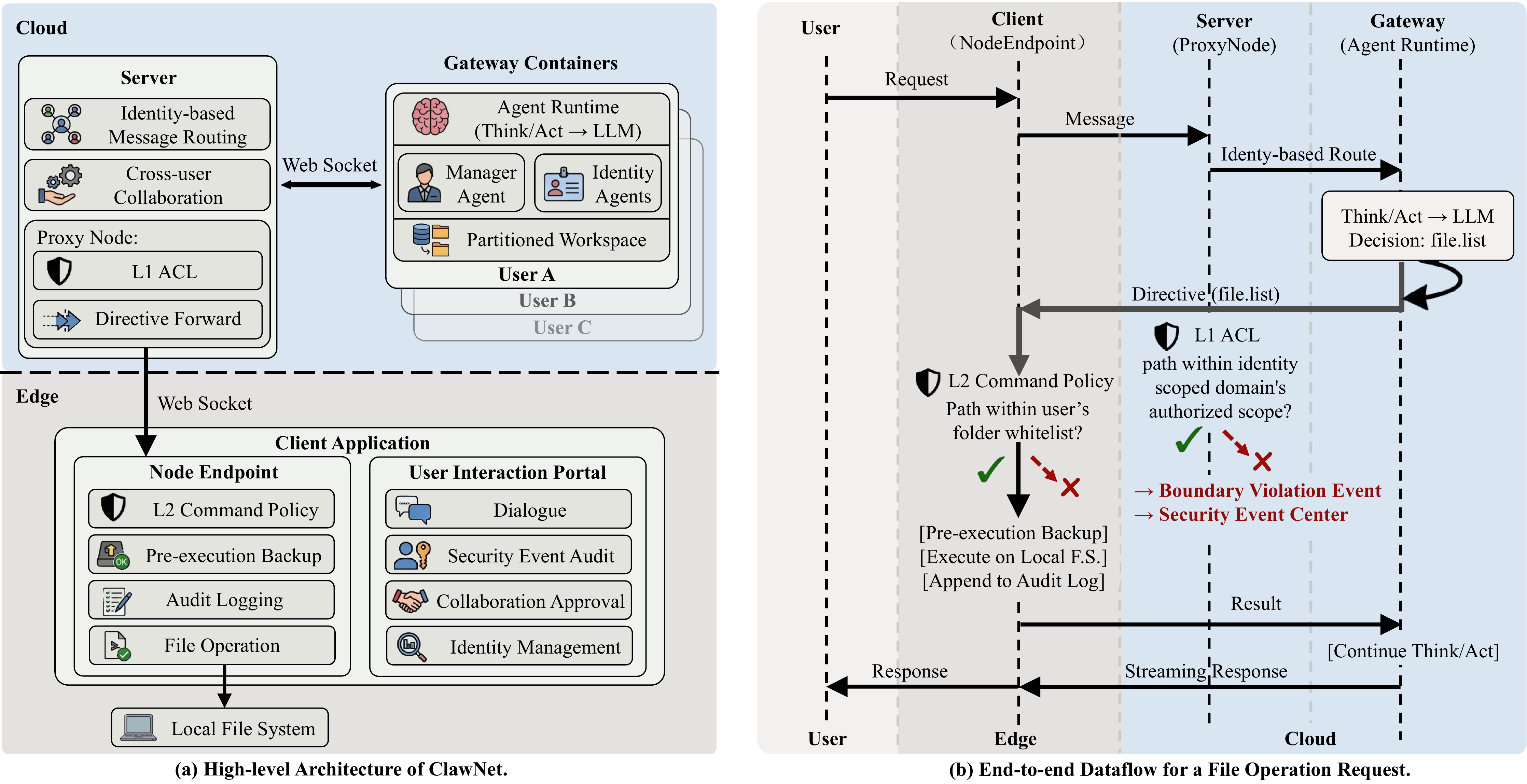}
  \caption{(a) High-level architecture of ClawNet. The cloud side 
  encompasses the server (central orchestration engine) and per-user 
  gateway containers (agent runtimes). The edge side consists of the 
  client application, which functions as both a user interaction 
  portal and a node endpoint for local file system operations. (b) End-to-end data flow for a file operation request. 
  Governance mechanisms, comprising identity binding, two-layer 
  scoped authorization (server-side L1 ACL and client-side L2 
  command policy), and accountability (pre-execution backup and 
  audit logging), are integrated directly into the directive 
  transmission path.}
  \label{fig:architecture}
\end{figure}

As illustrated in Figure~\ref{fig:architecture} (a), the cloud side 
comprises two primary components, including gateway containers and a central orchestration engine.

\textbf{Gateway containers.}
To ensure strict multi-tenancy isolation, each user is provisioned 
with a dedicated container running an agent execution engine with 
OS-level control capabilities. The agent runtime within the 
container employs a think/act loop to drive a large language model 
(LLM)-based reasoning, translating high-level natural language 
intent into discrete, actionable plans. Each container maintains an 
independent process space, network stack, and persistent storage, 
ensuring that the reasoning traces, memory states, and intermediate 
data of different users remains mutually inaccessible. $M_u$ and 
$\{I_u^1, \ldots, I_u^k\}$ reside within the same container to 
minimize communication overhead, as $M_u$ serves as a real-time 
internal advisor to identity agents. Despite co-residence, they 
operate in logically partitioned workspace subspaces corresponding 
to the authorization boundaries $\sigma_i$, preventing cross-domain 
data leakage (detailed in 
\S\ref{sec:identity-architecture}).

\textbf{Server (central orchestration engine).}
The server acts as the centralized backbone, responsible for 
identity-based message routing, cross-user collaboration 
orchestration (\S\ref{sec:identity-architecture}), and the 
first-layer access control list (L1 ACL) for file operations. It 
maintains persistent WebSocket connections to each user's gateway 
container. When an agent initiates a file operation, the server 
functions as a proxy node, a secure mediator between the cloud-side 
gateway and the edge-side client, verifying that 
$\text{target}(o) \in \sigma_i$ before forwarding directives to the 
user's device.

\subsection{Node Endpoint Overview}
\label{subsec:node-overview}

The edge side is embodied by the client application installed on 
the owner's device, serving two roles. First, it acts as a user 
interaction portal, providing the graphical interface for dialogue, 
identity management, collaboration approval, and security event 
auditing. Second, it acts as a \emph{node endpoint}, the terminal 
execution environment that receives operational directives from the 
cloud and performs physical input/output operations on the local 
file system. Upon initialization, the client registers with the 
server as a node, advertising its supported primitive command set. 
Figure~\ref{fig:architecture} (b) illustrates the end-to-end data flow of 
an operational directive.

\textbf{File operation capabilities.}
The node endpoint equips agents with a comprehensive suite of 
OS-level file primitives, encompassing reading, writing, directory 
traversal, relocation, renaming, replication, directory creation, 
and secure deletion. Operating within the think/act loop, agents 
autonomously orchestrate complex operation sequences to fulfill 
high-level user intent. The architecture maintains location 
transparency for the reasoning engine: the agent remains agnostic 
to the physical execution environment, merely issuing high-level 
directives to the node for local fulfillment. These capabilities 
are categorized into \texttt{read-only operations} (e.g., content 
retrieval, directory listing, and metadata queries) and 
\texttt{mutative operations} (e.g., writing, moving, and deletion). 
To ensure system integrity, every mutative action triggers an 
automated pre-execution backup, ensuring that any modification 
remains fully traceable and reversible.

\textbf{File access control.}
Extending OS-level file capabilities to agents inherently exposes 
the owner's most sensitive data to the agentic domain. ClawNet 
addresses this risk through a dual-layer independent authorization 
scheme, enforcing $\text{target}(o) \in \sigma_i$ along the entire 
directive path from cloud to edge. At the first layer 
(server-side), the server verifies the operation against the 
identity domain configuration of $I_u^i$. At the second layer 
(client-side), the client independently validates the operation 
against an explicit folder whitelist defined by the owner. The two 
layers function autonomously with a fail-closed design: a rejection 
by either layer results in a full-chain denial. The integration of 
this scheme with the layered identity architecture, including how 
different identity agents operate within disparate access scopes 
and how boundary violations are escalated, is detailed in 
Section~\ref{sec:identity-architecture}.
\begin{maintakeawaybox}
\begin{itemize}[nosep,leftmargin=1.5em]
    \item \textbf{Defense in depth, not defense in series.} The 
    two authorization layers operate independently. Even if the 
    cloud-side L1 ACL is compromised, the client-side L2 whitelist 
    still enforces the owner's boundary. Either layer alone can 
    deny an operation.
\end{itemize}
\end{maintakeawaybox}
\textbf{Operational auditing and reversibility.}
The node endpoint logs every mutative operation as 
$\ell = (o, u, I_u^i, \text{result}, t)$ into an append-only 
local audit log, with automatic pre-execution backup of all 
affected files. The owner may invoke single-step undo or batch 
rollback. File deletion is implemented as relocation to a 
recoverable staging area, and file overwrites automatically 
preserve the original content. These mechanisms secure the owner's 
sovereign right of correction over the agent's autonomous behavior.
\section{Identity Governance and Agent Collaboration}
\label{sec:identity-architecture}

To credibly collaborate on behalf of its owner, an agent must possess 
a stable and persistent identity. This section builds on the formal 
definitions in Section~\ref{sec:system-overview} to present the 
concrete mechanisms of identity-centric governance: how agent identity 
is established through memory-driven cognitive 
coupling~(\S\ref{subsec:identity-binding}), the layered architecture 
of the manager agent and identity agents along with their governance 
mechanisms~(\S\ref{subsec:layered-identity}), and the cross-user 
collaboration protocol built upon this identity governance 
framework~(\S\ref{subsec:cross-user-collaboration}).

\subsection{Identity Binding: From Task Memory to Cognitive Coupling}
\label{subsec:identity-binding}

Contemporary agentic systems have begun to incorporate long-term 
memory, but such capabilities typically focus on operational 
continuity, preserving user preferences to enhance future task 
efficiency. ClawNet redefines the role of memory: it serves not merely 
as a tool for task optimization, but as the foundational substrate for 
an agent to truly \emph{know} its owner.

\begin{maintakeawaybox}
\begin{itemize}[nosep,leftmargin=1.5em]
    \item \textbf{Identity binding is progressive, not configured.} 
    The agent builds owner understanding through three layers of 
    memory accumulation: factual memory, pattern memory, and value 
    memory. This progressive cognitive coupling is what transforms a 
    generic tool into a credible owner proxy 
    (Section~\ref{subsec:identity-binding}).
\end{itemize}
\end{maintakeawaybox}

Specifically, ClawNet's memory system achieves cognitive coupling 
through three progressive layers.

The first layer is factual memory. Through daily interactions, users 
delegate routine tasks via a natural language interface, including 
file organization, document archiving, information retrieval, and 
directory management. By executing these operations on the user's 
local file system via the node endpoint 
(\S\ref{subsec:node-overview}), the agent accumulates basic facts 
about its owner: locations of frequently used files, project directory 
structures, contact information, and schedules. These facts constitute 
the raw material of identity cognition.

The second layer is pattern memory. As interactions accumulate 
longitudinally, the agent extracts behavioral patterns from discrete 
facts: workflow habits (e.g., organizing emails every Monday, 
archiving financial files at month-end), file organization preferences 
(e.g., categorizing by project rather than by date), and 
context-dependent communication styles (e.g., concise and direct with 
colleagues, polite and formal with clients). Pattern memory enables 
the agent to anticipate the owner's intent rather than passively 
awaiting instructions.

The third layer is value memory. The agent further internalizes the 
owner's decision boundaries and value dispositions: which information 
must never be shared, what the bottom line is in negotiations, and 
whether the owner tends to compromise or hold firm when facing 
conflicts. This value-level understanding enables the agent, when 
acting as a proxy in cross-user collaboration, to know not only 
``what the owner would do'' but also ``what the owner would never do.''

ClawNet persists all three layers of cognition across sessions. The 
agent does not experience amnesia at the conclusion of a session but 
incrementally deepens its alignment with the owner through each 
successive engagement. Consequently, the relationship between the 
agent and its owner transcends technical isolation (dedicated 
containers and segregated workspaces) to achieve cognitive coupling: 
the agent forms a unique, irreplaceable personalized model of its 
owner. This cognitive coupling is the critical foundation for the 
agent's evolution from a generic utility into a sovereign 
representative.

\subsection{Layered Identity Architecture}
\label{subsec:layered-identity}

Once identity binding is established, a critical design question 
arises: should a user be represented by a single monolithic agent or 
a distributed set of specialized agents?

A monolithic agent suffers from an inherent tension: deep 
owner-alignment is essential for effective proxying, but it 
significantly elevates the risk of inadvertent privacy disclosure 
during external interactions. Conversely, constraining an agent's 
knowledge for security purposes inevitably compromises its capacity 
for cross-domain reasoning. ClawNet reconciles this conflict through 
a layered identity architecture, partitioning the user's agent system 
$\mathcal{A}_u = (M_u, \{I_u^1, \ldots, I_u^k\})$ into two 
distinct tiers.

\begin{maintakeawaybox}
\begin{itemize}[nosep,leftmargin=1.5em]
    \item \textbf{``Cannot'' beats ``should not.''} A monolithic 
    agent that knows everything about its owner is powerful but 
    dangerous: prompt-level constraints cannot reliably prevent 
    privacy leakage. The layered architecture resolves this by 
    letting $M_u$ hold complete cognition but remain architecturally 
    unexposed, while each $I_u^i$ faces the outside world carrying 
    only context-relevant knowledge. Security shifts from behavioral 
    reliability to structural determinism 
    (Section~\ref{subsec:layered-identity}).
\end{itemize}
\end{maintakeawaybox}
\textbf{Manager agent.}
$M_u$ serves as the authoritative anchor of the user's digital 
identity. It has the privilege to access the memory contents of all 
identity agents $\{I_u^1, \ldots, I_u^k\}$, enabling it to form a 
cross-domain global perspective. Crucially, $M_u$ does not maintain a 
separate global knowledge base; rather, it derives its global 
understanding through aggregated access to all identity agents' 
memories.

This global access capability makes $M_u$'s direct participation in 
external interactions inherently risky. Consider a scenario where 
$M_u$ can simultaneously access confidential commercial contract 
information stored in the ``work'' identity agent and private health 
records stored in the ``personal'' identity agent. If $M_u$ is 
queried with ``how have you been lately?'' during a collaborative 
session with a colleague, it might inadvertently leak health 
information in its response. Prompt-level security constraints such as 
``do not share private information'' rely on the model perfectly 
adhering to instructions across every inference step, a requirement 
that is statistically impossible to guarantee.

ClawNet therefore enforces architectural-level isolation on $M_u$: 
the collaboration protocol is structurally incapable of routing 
external messages to $M_u$. The security paradigm shifts from a 
policy-based ``should not'' (relying on model behavioral reliability) 
to an architectural ``cannot'' (leveraging system-level determinism).

Under this isolation constraint, $M_u$ assumes the role of an 
internal consultant. When $I_u^i$ requires cross-domain knowledge 
support during a collaboration, it may solicit guidance from $M_u$. 
$M_u$ provides advice based on its aggregated access to all identity 
agents' memories, subject to two constraints: $M_u$'s response is 
never sent directly to external collaborators but is instead 
presented to $I_u^i$ and the owner, who decide whether and how to 
integrate it; and $M_u$ is guided by $I_u^i$'s current context $c_i$ 
when formulating advice, avoiding unsolicited exposure of information 
from unrelated domains.

\textbf{Identity agents.}
Each $I_u^i = (c_i, \sigma_i, \mathcal{K}_i, \mathcal{P}_i)$ is a 
contextual projection of the owner within a specific collaborative 
environment. Individuals fulfill diverse social roles across different 
settings: a project manager in a professional context, a parent in 
family life, a researcher in academia. Each identity agent represents 
a digital instantiation of one such role, bound to an identity domain 
marked by a tag, with an independent knowledge space 
$\mathcal{K}_i \subset \mathcal{K}_u$ and persistent memory.

In implementation, different identity agents operate within logically 
isolated workspace subspaces inside the same gateway container. Each 
$I_u^i$ can only access files and resources within $\sigma_i$, and 
$I_u^i$ cannot read the workspace contents of $I_u^j$ ($i \neq j$), 
even though they belong to the same owner.

The core value of identity agents is realized in cross-user 
collaboration. The owner explicitly selects which identity to present 
when engaging with another user's agent: a ``work'' identity agent to 
coordinate project milestones with a colleague, or an ``academic'' 
identity agent to discuss research methodologies with a collaborator. 
The selected agent carries only the knowledge $\mathcal{K}_i$ of its 
identity domain, ensuring that information from other domains remains 
entirely inaccessible to the counterpart. The prerogative to select 
the active identity resides exclusively with the owner.

The lifecycle of an identity agent evolves in synchrony with the 
owner's social roles: instantiated when a new role emerges (with 
$\mathcal{P}_i$ and $\sigma_i$ configured), retired when the role 
concludes (with authorizations and records archived), and its 
parameters adjustable at any time.

\begin{maintakeawaybox}
\begin{itemize}[nosep,leftmargin=1.5em]
    \item \textbf{Complementary roles.} $M_u$ and $\{I_u^i\}$ form 
    a governed agency system. $M_u$ is the unified internal 
    cognition, able to access all identities' memories but never 
    externally exposed. Each $I_u^i$ is an outward-facing social 
    persona, specialized but controlled. Together they make ``deep 
    owner understanding'' and ``safe external collaboration'' no 
    longer contradictory 
    (Section~\ref{subsec:layered-identity}).
\end{itemize}
\end{maintakeawaybox}

\textbf{Identity-level authorization and accountability.}
The layered identity architecture provides clear anchors for both 
authorization and accountability. Each $I_u^i$'s file access scope is 
constrained by $\sigma_i$, enforced through the dual-layer 
authorization scheme described in 
Section~\ref{subsec:node-overview}. Authorization policies evolve 
dynamically with the owner's real-world social roles. The 
architectural isolation of $M_u$ constitutes an additional 
authorization boundary: it holds the most comprehensive access 
privileges but is structurally barred from external communication.

Every operation is bound to the originating identity agent's 
identifier and its identity domain, logged as 
$\ell = (o, u, I_u^i, \text{result}, t)$. The owner can precisely 
trace the provenance and authorization basis of any action. 
Unauthorized access attempts are automatically intercepted and 
dispatched as boundary violation events to the owner's security event 
center, ensuring that every breach is audited and cannot be ignored.

\subsection{Cross-User Collaboration}
\label{subsec:cross-user-collaboration}

The preceding sections described identity governance within a single 
user domain. This section extends the scope to the inter-user 
dimension: the mechanisms governing collaboration between identity 
agents belonging to different owners.

\begin{maintakeawaybox}
\begin{itemize}[nosep,leftmargin=1.5em]
    \item \textbf{Agents automate the process, not the decision.} 
    In cross-user collaboration, agents handle information exchange, 
    solution deliberation, and progress synchronization. But the 
    authority to initiate collaboration, select the active identity, 
    delineate information-sharing boundaries, and make final 
    decisions remains with the human owner 
    (Section~\ref{subsec:cross-user-collaboration}).
\end{itemize}
\end{maintakeawaybox}

\textbf{Establishing contact relationships and identity assignment.}
The prerequisite for collaboration is a contact relationship between 
two users. A user sends a friend request; upon confirmation, they 
become contacts. The user then designates which identity agent to 
present to each contact: a ``work'' identity agent for a colleague, 
an ``academic'' identity agent for a collaborating advisor. When 
facing different counterparts, the same user presents distinct 
identity projections, each carrying its own $\mathcal{K}_i$ and 
$\sigma_i$. Users can dynamically adjust identity assignments or 
remove contacts at any time. The construction of the collaboration 
network is entirely in the hands of the user.

\textbf{Collaboration initiation and bilateral approval.}
A collaborative interaction is triggered when $I_u^i$ manifests the 
intent to communicate with another user during a local conversation, 
for instance, determining the need to retrieve information from a 
colleague. The system does not autonomously initiate the 
collaboration. Instead, it intercepts the action and issues an 
authorization request to the owner. The request is dispatched to the 
counterpart only after securing the owner's explicit approval. The 
counterpart's owner must also approve. A collaboration session is 
established only when both parties provide explicit consent:
$$\text{approve}(u, I_u^i) \wedge 
\text{approve}(v, I_v^j) \wedge 
u \in \mathcal{P}_j \wedge v \in \mathcal{P}_i$$
A rejection by either party immediately terminates the initiation 
sequence. Without this mandatory bilateral approval, an agent might 
unilaterally share information without the owner's awareness, an 
outcome that is fundamentally unacceptable when the interacting 
agents represent distinct owners with independent interests.

\textbf{Multi-turn dialogue and autonomous collaboration.}
Once the session is established, the identity agents of both parties 
enter an autonomous multi-turn dialogue. Agents exchange information, 
deliberate on solutions, and orchestrate tasks, each generating 
responses strictly based on its own $\mathcal{K}_i$. At every turn, 
the server injects role-specific system prompts that clearly define 
each agent's identity, stance, and behavioral norms, preventing the 
well-known problem of role drift in multi-turn LLM conversations. The 
server simultaneously performs identity verification and authorization 
checks.

Conversation termination is governed by three layers of control. At 
the agent layer, the agent evaluates dialogue progression through 
semantic state markers and gracefully concludes the conversation upon 
achieving its objective. At the system layer, a maximum turn threshold 
prevents indefinite continuation. At the owner layer, the human own
\section{Related Work}
\label{sec:related}

\paragraph{Single-agent and multi-agent frameworks.}
Single-agent systems such as OpenClaw~\citep{openclaw2025} and 
Anthropic's Computer Use~\citep{anthropic2024computeruse} have 
advanced agent capability from tool augmentation to OS-level 
control, but address only what one agent can do for one user. 
Multi-agent frameworks such as MetaGPT~\citep{hong2023metagpt}, 
AutoGen~\citep{wu2023autogen}, CrewAI~\citep{crewai2024}, 
ChatDev~\citep{qian2024chatdev}, and 
CAMEL~\citep{li2023camel} demonstrate effective collaboration 
among multiple agents through role-playing and structured 
workflows. However, all operate within a single principal's 
scope: agents share the same user, the same goal, and the same 
trust domain. When MetaGPT assigns agents the roles of 
``Product Manager'' and ``Engineer,'' these are functional 
specializations, not representatives of different humans with 
divergent interests. ClawNet addresses a fundamentally different 
setting: each agent is permanently bound to a distinct human 
owner, and cross-user cooperation takes the form of governed 
inter-agent negotiation.

\paragraph{Agent safety and governance.}
A growing body of work recognizes the governance gap in agentic 
systems. \citep{shavit2023governing} outline practices for 
governing agentic AI but do not provide a system implementation. 
\citep{chan2025infrastructure} identify the need for 
infrastructure supporting agent identity and interaction, and 
\citep{south2025authenticated} propose authenticated delegation 
as a prerequisite for authorized AI agents. 
\citep{casper2025agentindex} catalog the landscape of AI agents 
and their accountability gaps, while 
\citep{schroeder2025challenges} discuss the structural 
challenges of multi-agent security. These works collectively 
establish the theoretical motivation for governance-aware agent 
systems but stop short of an end-to-end implementation. ClawNet 
translates these identified requirements into a deployed system 
with concrete enforcement mechanisms.

\paragraph{Agent interoperability protocols.}
Standardized agent communication has a long history, from 
KQML~\citep{finin1994kqml} to FIPA Agent Communication Language (FIPA-ACL)~\citep{pitt1999some}. More recently, Google's Agent2Agent protocol~\citep{google2025a2a} provides a 
communication layer for agents to discover and exchange messages 
across framework boundaries, and Anthropic's Model Context 
Protocol enables standardized tool integration. These protocols 
address \emph{how} agents communicate but not \emph{under what 
governance}: they do not bind agents to specific owners, enforce 
authorization scopes, or produce auditable responsibility chains. 
ClawNet can be viewed as a governance layer that operates on top 
of such communication protocols, adding identity binding, scoped 
authorization, and action-level accountability to the 
interaction substrate.

\section{Conclusion}
\label{sec:conclusion}

We have presented ClawNet, an identity-governed agent collaboration 
framework that instantiates a human-symbiotic paradigm for cross-user 
autonomous cooperation. By permanently binding each agent to a 
specific human owner through progressive cognitive coupling, 
partitioning identity into privacy-preserving and outward-facing 
tiers, and enforcing identity binding, scoped authorization, and 
action-level accountability throughout every interaction, ClawNet 
moves beyond the digitization of individual skills toward the 
digitization of human collaborative relationships themselves.

We believe this work points to a broader shift in how AI agents 
should be designed. As agents grow more capable and autonomous, they 
will inevitably need to interact with agents belonging to other 
people, organizations, and institutions. The governance 
infrastructure for such interactions cannot be an afterthought. Just 
as the internet required protocols for identity and trust before 
e-commerce could flourish, the emerging ecosystem of autonomous 
agents requires governance primitives before cross-user agent 
collaboration can become a practical reality. We hope that ClawNet 
serves as both a working prototype and an open invitation for the 
community to build upon this foundation.

\bibliography{ref}
\bibliographystyle{icml2026}

\end{document}